\documentclass[conference]{template/IEEEtran}   
\makeatletter
\def\ps@IEEEtitlepagestyle{%
  \def\@oddfoot{\mycopyrightnotice}%
  \def\@evenfoot{}%
}
\def\mycopyrightnotice{%
  {\footnotesize U.S. Government work not protected by U.S. copyright\hfill}
  \gdef\mycopyrightnotice{}
}

\IEEEoverridecommandlockouts                              

\usepackage{cite}
\usepackage{balance}
\usepackage{amsmath,amssymb,amsfonts}
\usepackage{algorithmic}
\usepackage{graphicx}
\usepackage{textcomp}

\usepackage[utf8]{inputenc}
\usepackage[T1]{fontenc}
\usepackage{mathptmx} 
\usepackage{mathptmx} 
\usepackage{amsmath} 
\usepackage{amssymb}  
\usepackage{graphicx} 
\usepackage{grffile}
\usepackage{array}
\usepackage{mathtools}
\usepackage{commath}
\usepackage{colortbl}
\usepackage{enumerate}
\usepackage{hyperref}
\usepackage{cleveref}
\usepackage{subfig}
\usepackage{float}
\usepackage{tabularx}
\usepackage{makecell}
\usepackage{multirow}
\usepackage{placeins}
\usepackage{hyperref}
\usepackage{tabu}

\usepackage{blindtext}
\usepackage{eso-pic}
\IEEEoverridecommandlockouts
\usepackage{cite}
\usepackage{amsmath,amssymb,amsfonts}
\usepackage{algorithmic}
\usepackage{graphicx}
\usepackage{textcomp}
\usepackage{xcolor}
\def\BibTeX{{\rm B\kern-.05em{\sc i\kern-.025em b}\kern-.08em
    T\kern-.1667em\lower.7ex\hbox{E}\kern-.125emX}}
    
\usepackage{eso-pic}
\newcommand\AtPageUpperMyright[1]{\AtPageUpperLeft{%
 \put(\LenToUnit{0.17\paperwidth},\LenToUnit{-2cm}){%
     \parbox{0.9\textwidth}{\raggedleft\fontsize{8}{11}\selectfont #1}}%
 }}%
\newcommand{\conf}[1]{%
\AddToShipoutPictureBG*{%
\AtPageUpperMyright{#1}
}
}  

\begin{document}
\title{\vspace*{1cm} Optimizing Lead Time in Fall Detection for a Planar Bipedal Robot\\
}

\author{\IEEEauthorblockN{M. Eva Mungai}
\IEEEauthorblockA{\textit{Mechanical Engineering} \\
\textit{University of Michigan}\\
Ann Arbor, MI US \\
mungam@umich.edu}
\and
\IEEEauthorblockN{Jessy Grizzle}
\IEEEauthorblockA{\textit{Department of Robotics} \\
\textit{University of Michigan}\\
Ann Arbor, MI US \\
grizzle@umich.edu}}

\maketitle
\conf{\textit{  Proc. of the International Conference on Electrical, Computer, Communications and Mechatronics Engineering (ICECCME 2023) \\ 
19-21 July 2023, Tenerife, Canary Islands, Spain}}



\begin{abstract}
For legged robots to operate in complex terrains, they must be robust to the disturbances and uncertainties they encounter. This paper contributes to enhancing robustness by designing fall detection/prediction algorithms that will provide sufficient lead time for corrective motions to be taken. Falls can be caused by abrupt (fast-acting), incipient (slow-acting), or intermittent (non-continuous) faults. Early fall detection is a challenging task due to the masking effects of controllers (through their disturbance attenuation actions), the direct relationship between lead time and false positive rates, and the temporal behavior of the faults/underlying factors. In this paper, we propose a fall detection algorithm capable of detecting both incipient and abrupt faults while maximizing lead time and meeting desired thresholds on the false positive and negative rates.
\end{abstract}

\section{Introduction}\label{sec:introduction}
\subsection{Motivation}
For legged robots to successfully navigate in the real world, it's imperative to employ methods to predict the potential occurrence of a fall and, if possible, execute reflexive motions to either prevent the fall from happening or to make the ``landing'' less dangerous, to the robot or its surroundings. The infeasibility of accounting for all potential perturbations and uncertainties while operating autonomously in dynamic environments makes falling almost inevitable. Here, we focus on bipedal robots.

Fall detection and recovery algorithms often consist of two parts: fall detection and reflexive motions. The focus of this paper, however, is solely on fall detection. The objective is to reliably predict potential falls of bipedal robots with sufficient time to deploy a recovery strategy. Bipedal robots are chosen because their smaller support polygon causes them to be inherently less stable in comparison to robots with more legs. To simplify the fall detection problem while providing a pathway to scale up to more complex dynamic motions and robots, the task of standing with a planar four-link biped is chosen. As our task of interest is standing, we define a ``fall' as any link other than the feet coming in contact with the ground  \cite{ref:pratt2006velocity} or if the feet are off the ground. 

\subsection{Background}
In the following, the term falls will be associated with the term faults, which are defined as unacceptable deviations from expected behavior in at least one variable \cite{ref:ISERMANN19827}. Faults can be classified based on their time dependency, location, and underlying causal factors. Temporally, faults can either be abrupt (step-like or rapidly varying), incipient (drift-like or slowly varying), or intermittent (non-continuous). Faults can be caused by either internal or external disturbances and uncertainties \cite{ref:ISERMANN19827,ref:safaeipour2021survey,ref:renner2006instability}. Incipient faults arise from a gradual deviation from the robot's nominal states or assumed environment, while, as the name suggests, abrupt faults typically arise from shoves or unexpected impacts with the environment. Intermittent faults are non-continuous faults. We do not address intermittent faults in this paper.

In general, the states of the robot can be divided into three classes: safe/balanced, falling, and fallen \cite{ref:kalyanakrishnan2011learning}. The safe/balanced states are states where it is possible for the robot to avoid falling while under the influence of its nominal feedback controller. These states are therefore contained in a subset of the viability kernel \cite{ref:wieber2002stability,ref:hohn2009probabilistic,ref:wu2021falling}.

\subsection{Literature Review} 
A fall detection algorithm has been implemented in the commercial bipedal robot, Digit \cite{ref:ar_digit}. However, this does not appear to be the norm. The objective of most bipedal fall detection algorithms found in literature is to reliably and promptly detect all abrupt faults. Incipient faults are not considered.

Fault detection reliability has been determined using a combination of evaluation terms from the confusion matrix, such as false positive and negative rates \cite{ref:renner2006instability, ref:kalyanakrishnan2011learning,  ref:wu2021falling, ref:subburaman, ref:kim, ref:hohn2009probabilistic, ref:gallego2010continuous,ref:marcolino2013detecting, ref:ruiz2010fall, ref:karssen2009fall}. Thresholds, based on factors such as center of mass height, have been proposed to minimize the percentage of false negative fault declarations in \cite{ref:kalyanakrishnan2011learning}, while the output of the fall detection algorithm is monitored for a certain number of windows ($N_{monitor}$) to reduce the false positive rate in \cite{ref:kalyanakrishnan2011learning,ref:subburaman,ref:khalastchi2015online}. Lead time, defined as the difference between the time of the actual fall and the predicted fall, is used to inform whether or not sufficient time is left for the implementation of recovery/reflexive motions. It is desirable to have a large lead time, however, maximizing lead time can increase false positive rates. The minimum amount of lead time needed depends on the chosen recovery algorithm and the robot's dynamics, as discussed in \cite{ref:renner2006instability,ref:kalyanakrishnan2011learning,ref:wu2021falling,ref:suetani2011nonlinear}

Fall detection algorithms can either rely on physics-based \cite{ref:amri2018improved,ref:muender,ref:xinjilefu}  or data-based models \cite{ref:kalyanakrishnan2011learning, ref:wu2021falling, ref:subburaman, ref:hohn2009probabilistic, ref:ruiz2010fall,ref:karssen2009fall, ref:marcolino2013detecting}. Physics-based models can suffer from model inaccuracies while data-based models are limited by the amount of data available. For both physics-based and data-based models, the objective is to obtain either a model of the nominal (safe) states and/or of the faulty (unsafe) states. However, this is not simple. In practice, it is infeasible to quantify all faults that can lead to a fall, and the faulty states in any given trajectory are irregular and rare. Both of these conditions make it nearly impossible to obtain an accurate model of the anomalies. It is also challenging to obtain a model that accounts for all the safe states of the robot. However, due to advances in the machine learning community, data-based algorithms are becoming more common.

Data-driven detection algorithms can be divided into two subsequent parts, feature engineering and the method used for detection. Feature engineering consists of selecting and transforming raw data into features that can differentiate between faulty and normal states.  Even though stability metrics are used to increase the robustness of controllers, they individually are not a sufficient condition for falling \cite{ref:subburaman}. A combination of stability metrics from bipedal control theory, such as the angular momentum about the center of mass ($L_{com}$), and kinematic functions, such as the center of mass position ($p_{com}$), are typically chosen as features \cite{ref:kalyanakrishnan2011learning, ref:wu2021falling, ref:subburaman, ref:hohn2009probabilistic, ref:ruiz2010fall, ref:karssen2009fall, ref:marcolino2013detecting}.

Classification algorithms, such as that used by \cite{ref:kalyanakrishnan2011learning}, attempt to learn a model from labeled training data and then classify a data point into one of the classes based on the learned model. A disadvantage of classification algorithms is that they can output incorrect predictions if the input data is outside the training data parameters (outside distribution). Nearest-neighbor-based algorithms, such as \cite{ref:khalastchi2015online}, assume that normal data exist in highly dense spaces whereas the neighborhood of anomalous data is sparse. However, these algorithms can have high false positives if the normal instances do not exist in sufficiently dense neighborhoods. Threshold-based algorithms, such as \cite{ref:subburaman}, attempt to use a combination of features to derive a threshold that can be used to separate faulty and normal states.

\subsection{Objective of the Paper}

\textbf{For the task of standing} and for given \textbf{upper bounds on the false positive and false negative rates}, our objective is to detect potential falls caused by either \textbf{incipient or abrupt faults} while \textbf{maximizing the lead time}, that is, the time from fault declaration to the robot entering a fallen state. The objective is challenging due to the crowding phenomenon \cite{ref:safaeipour2021survey,ref:Chen2020}, masking effects of the controller as it tries to mitigate deviations from steady state \cite{ref:ISERMANN19827}, and the direct relationship between lead time and false positive rate. The crowding phenomenon is the similarity between the normal and incipient faulty data which makes it difficult to separate normal data from faulty data \cite{ref:safaeipour2021survey,ref:Chen2020}.

To achieve this objective, we design a nearest-neighbor detection algorithm and compare its performance to an existing classification-based detection algorithm. A threshold-based method was not chosen for comparison because it is difficult to find simple thresholds for systems as complex as bipedal robots.

\subsection{Contributions} 
The major contributions of the paper are as follows:
\begin{itemize}
    \item An algorithm that maximizes lead time subject to bounds on false positive and negative rates;
    \item A method of identifying trajectories associated with incipient or abrupt faults;
    \item A way to label the data based on lead time is proposed;
    \item A nearest-neighbor classification-based fall detection algorithm that can detect both incipient and abrupt faults; and    
        \item A comparison of the proposed nearest-neighbor classification algorithm and an existing classification algorithm.
\end{itemize}

\section{Robot Description and Data Generation}\label{sec:robot_description_data_generation}
In this section, we describe the robot model that is used for this study and how the data are generated and prepared.

\subsection{Robot Description}
\begin{figure}[!h]
    \centering
    \includegraphics[width=0.23\textwidth]{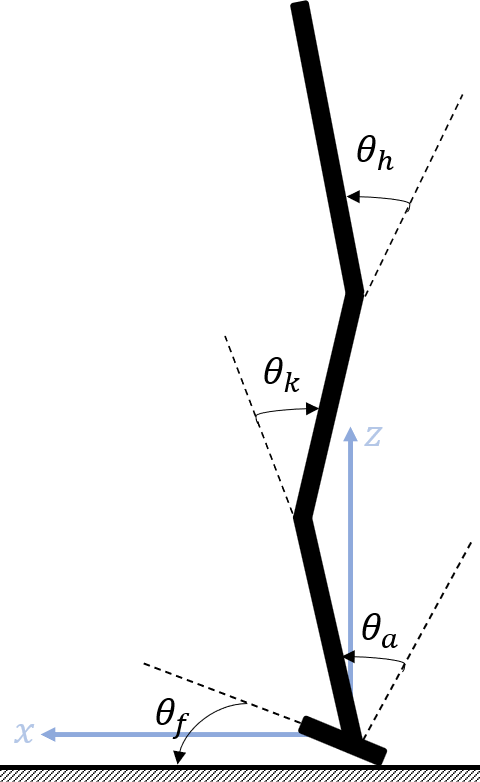}
    \captionsetup{font=scriptsize}
    \caption{Four-link robot model used in our study.
    }
    \label{fig:fourlinkrobot}
\end{figure}

We assume a four-link planar robot based on Wandercraft's exoskeleton Atalante \cite{ref:wandercraft, ref:omar}. The four links are joined by three actuated revolute joints, called the ankle, knee, and hip. Fig.~\ref{fig:fourlinkrobot} depicts the chosen robot.  

\subsection{Equations of Motion and Simulation Environment}
The equations of motion are given by \eqref{eq:floating_base_fall} and \eqref{eq:accelConstEq} 
\begin{align}
    \label{eq:floating_base_fall}
    D(q)\ddot{q}+C(q,\dot{q})\dot{q}+G(q)&= Bu+J^T(q)\Gamma \\
    \label{eq:accelConstEq}
    J(q)\ddot{q} + \dot{J}(q,\dot{q})\dot{q}&=0,
\end{align}
where $q$ is the vector of generalized coordinates defined by \eqref{eq:gen_coords}, $u$ is the torque input vector, $D$, $C$, $G$, and $B$ are the inertia, Coriolis, gravity, and torque distribution matrices/vector, respectively, $J$ is the Jacobian mapping the contact wrenches to the generalized coordinates, and $\Gamma$ is the contact wrench. The floating-base Lagragian model is given by \eqref{eq:floating_base_fall} while the contact/acceleration constraint is given by \eqref{eq:accelConstEq} \cite{ref:Jessy,ref:mlsTxtBook,ref:lynch2017modern}. 
\begin{equation}
   q = \begin{bmatrix}
     \text{\rm foot~x}  \\
    \text{\rm foot~z} \\
    \text{\rm foot~angle ($\theta _f$)} \\
    \text{\rm ankle~angle  ($\theta _a$)} \\
    \text{\rm knee~angle  ($\theta _k$)} \\
    \text{\rm hip~angle  ($\theta _h$)} 
    \end{bmatrix}.
    \label{eq:gen_coords}
\end{equation}

A PD controller to maintain the robot in a standing position was designed and implemented in MATLAB \cite{ref:MATLAB}. The PD controller seeks to keep the foot flat on the ground while maintaining the center of mass (CoM) inside the support polygon.  The simulation environment uses MATLAB's ODE45 function and compliant ground contact forces represented as a spring-damper. 

\subsection{Data Generation } 
Four hundred trajectories each are generated for abrupt and incipient faults, with a sampling time of 0.03s and a disturbing force applied to the torso. To emulate disturbances that might cause the robot to oscillate slightly while standing, a random impulse force in the range of 0-159N and lasting for 0.075s is applied at time zero, but only the data after 2 seconds is kept. The abrupt faults last for 0.075 seconds with magnitudes ranging from 0-320N , while the incipient faults last for 1.0 seconds and range from 0-46N. The ranges, based on previous experiments, are chosen such that half the trajectories end in a fall (we'll refer to these trajectories as \textit{faulty trajectories}), and half of the safe (non-falling) trajectories have a heel or toe lift. Similar to \cite{ref:kalyanakrishnan2011learning}, the force magnitudes are generated using a uniform distribution. The abrupt force is applied at random between 2.5s and 3.5s, while the incipient fault is applied at random between 2.0s and 3.5s. The application time for the incipient fault is longer because, in order not to include an abrupt deviation in the robot's nominal states, only the data collected after the force is applied is kept.

\subsection{Data Pre-processing}\label{sec:data_preparation}
The features are selected as
\begin{equation}
\begin{bmatrix}
    L_{cop} - L_{com}\\
    p_{com}^x\\
   v_{com}^x\\
   (p_{toe} - p_{com})^x\\
    (p_{heel} - p_{toe})^{xz} \\
    (L_{cop} + L_{com})* sgn(p_{com}^x- p_{f_{mid}}^x) 
\end{bmatrix}
    \label{eq:features_main}
\end{equation}
where $v_{com}$ is the CoM velocity,  $p_{toe}$, $p_{heel}$,  and $ p_{f_{mid}}$ are the position of the toe, heel and middle of the foot, and $L_{cop}$ is the angular momentum about the contact point\footnote{The contact point is set to the rotation point (toe or heel) when the foot rotates, and the center of pressure otherwise.}. These features are chosen based on their correlation with the lead time and other features commonly used in literature. The distance correlation coefficient is used to evaluate the correlations as it is able to capture nonlinear relationships \cite{ref:szekely2014partial}.

The features are split into training (60\%), validation (20\%), and testing (20\%) sets using  scikit-learn's \cite{ref:scikit-learn} stratified train\_test\_split method and k-folds methods with the number of folds set to 5. The stratified methods are chosen because they ensure that each of the splits has the same distribution of normal and faulty data. Scikit-learn's min-max scaler is used to scale the data to a range of $\{0,1\}$. To ensure that only transient data is kept for training, only the first 6s of trajectories that are deemed as normal are kept.

\section{Fall Detection Methods}
As a baseline, we use the SVM-based classification algorithm of \cite{ref:scikit-learn}, while the nearest-neighbor classification algorithm is based on the Ward minimum variance method \cite{ref:ward1963hierarchical}. To prioritize recent data points over previous ones, both methods make use of sliding windows. The number of data points in a window is referred to as $N_{window}$. It is important to note that both methods are supervised algorithms.

\subsection{SVM Classifier}

The radial basis function is chosen as the kernel and the soft margin formulation is implemented for the SVM classifier (SVM classification algorithm). The training data for the classifier is defined as 
\begin{align*}
D&=\{X_i,y_i\}_{i=1}^n \\
&~~\text{where}  \\
&\begin{array}{rcl}
    n &=& \text{number of windows across all training data}\\
    m &=& \text{number of time steps in a window}\\
    x_{ij}&=&\text{features at time step j in window i} \\
        X_i &=& \begin{bmatrix} x_{i1} && x_{i2} && \cdots && x_{im}  \end{bmatrix}^\intercal \\
    y_i &\in&\  \left\{
    \begin{array}{rcl}
           -1  &\quad& X_i ~\in ~\text{faulty trajectory} \\
            1  &\quad& X_i ~\in ~\text{normal trajectory}
    \end{array}
    \right\}
    \end{array}
\end{align*}

\subsection{Nearest-Neighbor Classification Algorithm}
\label{sec:classifier_intro}
 The selected nearest-neighbor classification algorithm determines distance using the Ward minimum variance and a weighted Euclidean distance. Given two clusters A and B, the Ward minimum variance method calculates the effort, $E_{AB}$, it takes to join the two clusters together, as determined by the sum of squared errors, specifically,  
 \begin{equation}
    \label{eq:ward_method}
    E_{AB} = SSE_{AB} - SSE_{A} - SSE_{B},
    \end{equation} 
 where
\begin{align*}
    &SSE_{AB} =(A\cup B- \mu_{A\cup B})^\intercal R^{-1}_{A\cup B}(A\cup B- \mu_{A\cup B})\\
    &SSE_{A} = (A- \mu_{A})^\intercal R^{-1}_{A}(A- \mu_{A})\\
    &SSE_{B} =  (B- \mu_{B})^\intercal R^{-1}_{B}(B- \mu_{B})\\
    &R = \text{correlation coefficient matrix}\\
    &\mu = \text{mean vector}
\end{align*}

In our application, cluster B contains the features at the current time step while cluster A contains all the features in the previous time steps included in the window. Given, that B is a single data point, the Ward minimum variance simplifies to
\begin{equation}
    \label{eq:our_ward_method}
    E_{AB} = SSE_{AB} - SSE_{A}. 
\end{equation}

 The nearest-neighbor classification algorithm detects a potential fall if the effort it takes to join the two clusters A and B is higher than a threshold determined from the training data. The threshold is calculated offline as the maximum $E_{AB}$ value for the safe data while $R$ is determined using distance correlation. Note that the underlying assumption for the nearest-neighbor classification algorithm is that cluster A only contains safe data points. 

\subsection{Data Labeling}
As one of our objectives is to maximize the lead time, we propose the use of a training lead time to label windows in a trajectory. Training lead time is defined as the difference between the time of the actual fall and the time when a  sliding window of a trajectory can be labeled as faulty. Therefore, training lead time is a subset of the maximum lead time that can be achieved in a faulty trajectory. While labeling all windows in a faulty trajectory as faulty would achieve the maximum lead time, it would also increase the rate of false positives. For instance, given two faults that are close in magnitude but where one results in a fall and the other is safe, the safe trajectory could be mistaken as a faulty trajectory. 

 If a trajectory does not contain a fall, all windows derived from the trajectory are labeled as 1. If a trajectory ends in a fall, all windows containing data points after the desired training lead time are labeled as -1. Note that for an abrupt fault, the training lead time is only defined after the push is introduced, and only the data points before a fall are kept for the training data of both faults.

 The desired training lead time is determined by a grid search algorithm that trains the algorithm of interest using a range of training lead times from 0 to 2s and evaluates the results on the training and/or validation data. The training data is included in the evaluation process for cases where the algorithm is allowed to make mistakes, such as when using a soft margin in SVM. A training lead time of 2s would label all the data points in a faulty trajectory as faulty.

\subsection{Performance of Fall Detection Methods} 
\label{sec:fault_comparison}

In this section, we analyze the performance of the proposed nearest-neighbor classification algorithm and the SVM classifier. The algorithms are trained and evaluated on testing data across all 5 folds using just abrupt trajectories, just incipient trajectories, and both trajectories together. The evaluation metrics are false positive and negative rates, and the average lead time achieved. The desired false positive and false negative rates are set to 0. The training lead time chosen is the maximum that meets the given bounds on the false positive and false negative rates when evaluated on the training and validation data. Based on previous experiments, we set the values of the remaining hyper-parameters as $N_{window}=10$ and $N_{monitor} = 1$. 

\begin{figure*}
\subfloat[Nearest-Neighbor Incipient Fault]{\resizebox{.40\linewidth}{!}{\includegraphics[]{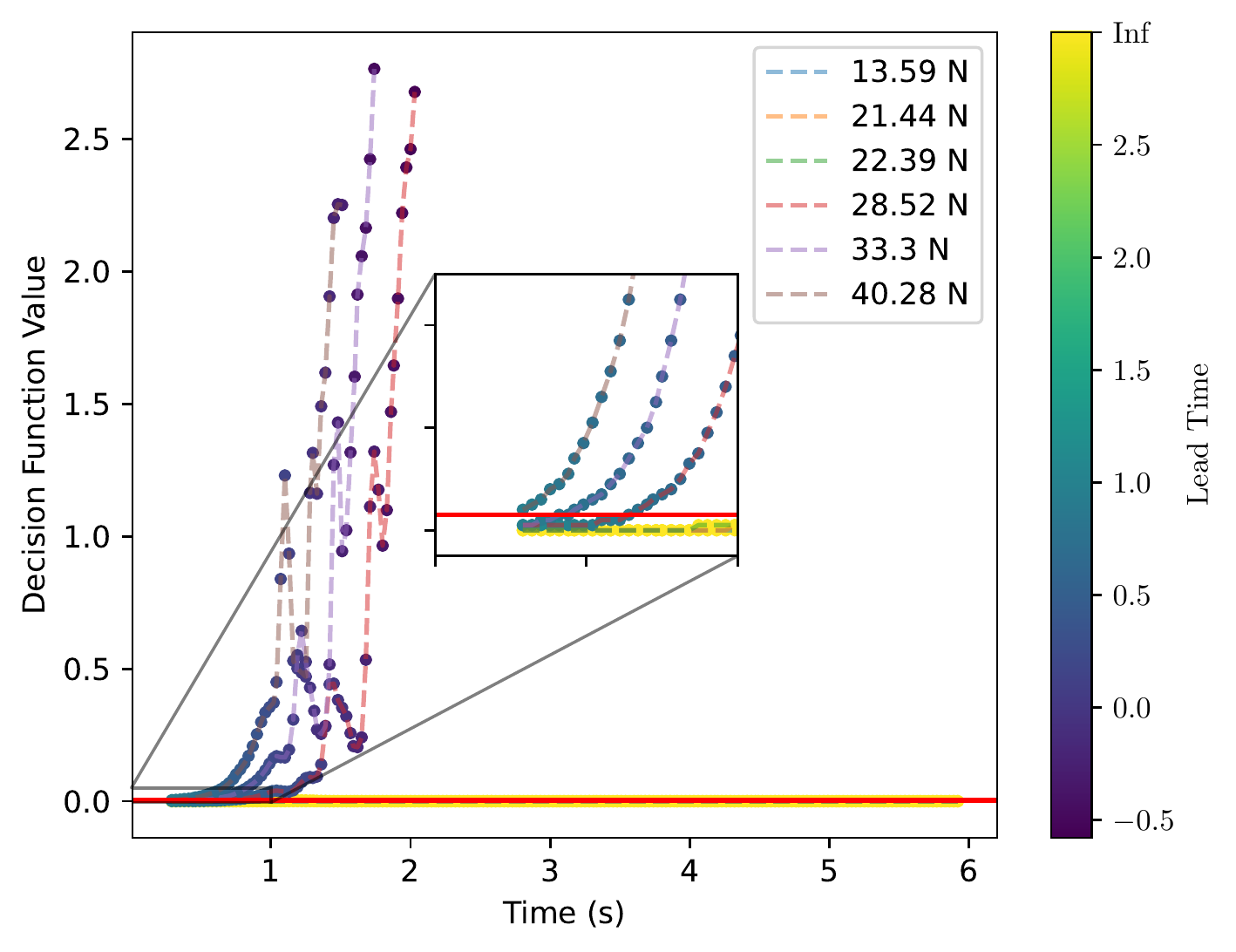}}}
\hfill
\subfloat[SVM Incipient Fault]{\resizebox{.40\linewidth}{!}{\includegraphics[]{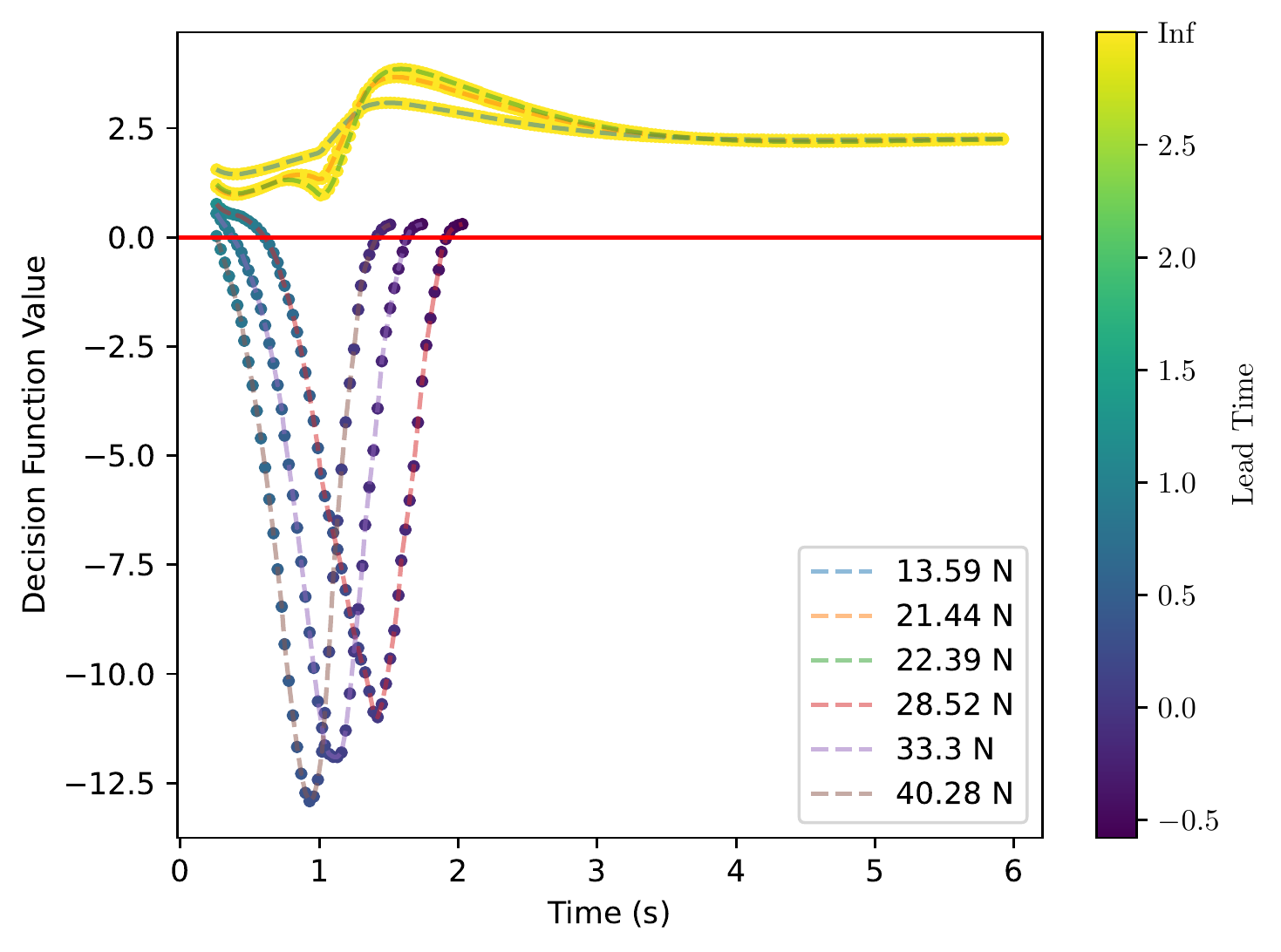}}}
    \hfil
    \newline
\subfloat[Nearest-Neighbor Abrupt Fault]{\resizebox{.40\linewidth}{!}{\includegraphics[]{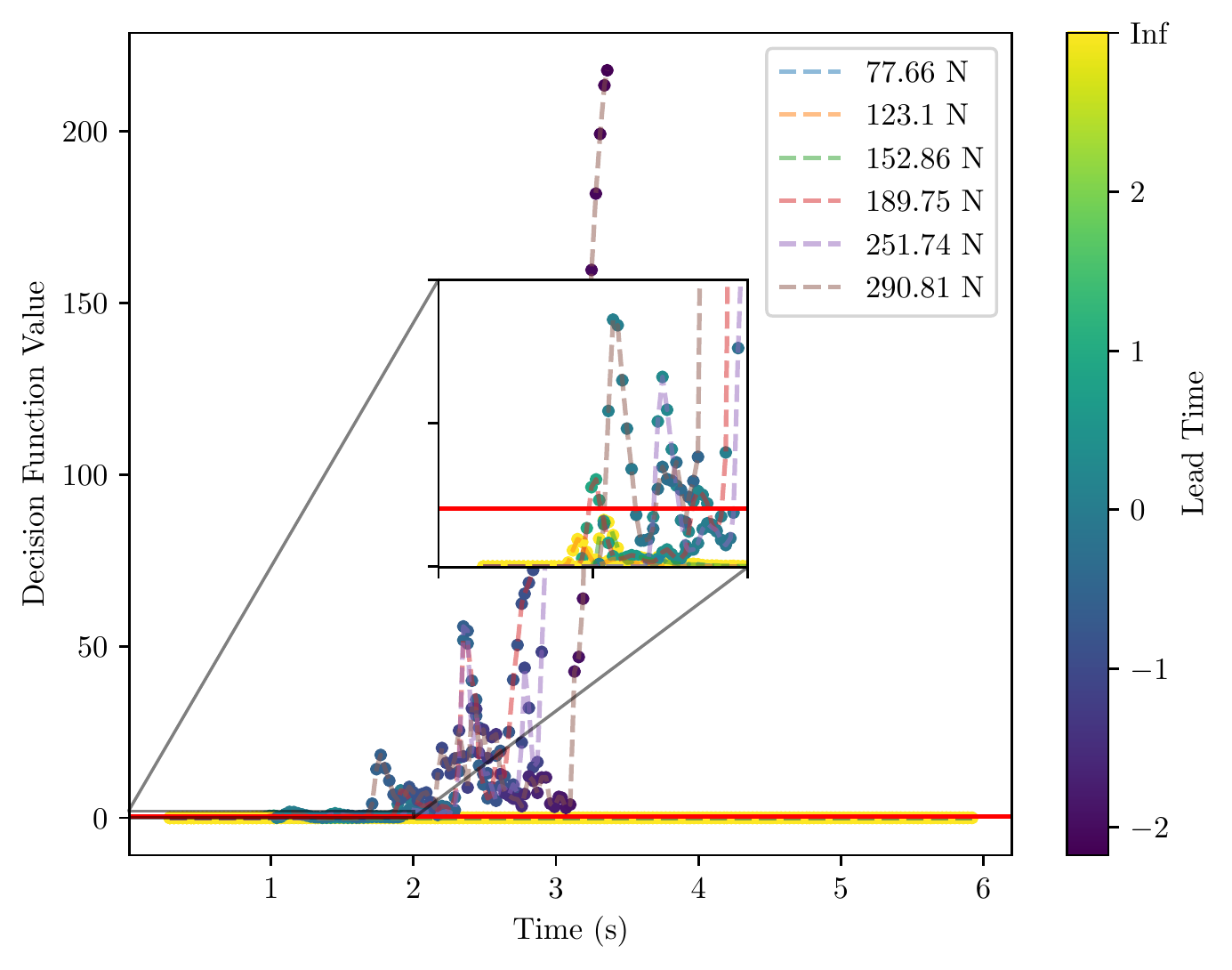}}}
\hfill
\subfloat[SVM Abrupt Fault]{\resizebox{.40\linewidth}{!}{\includegraphics[]{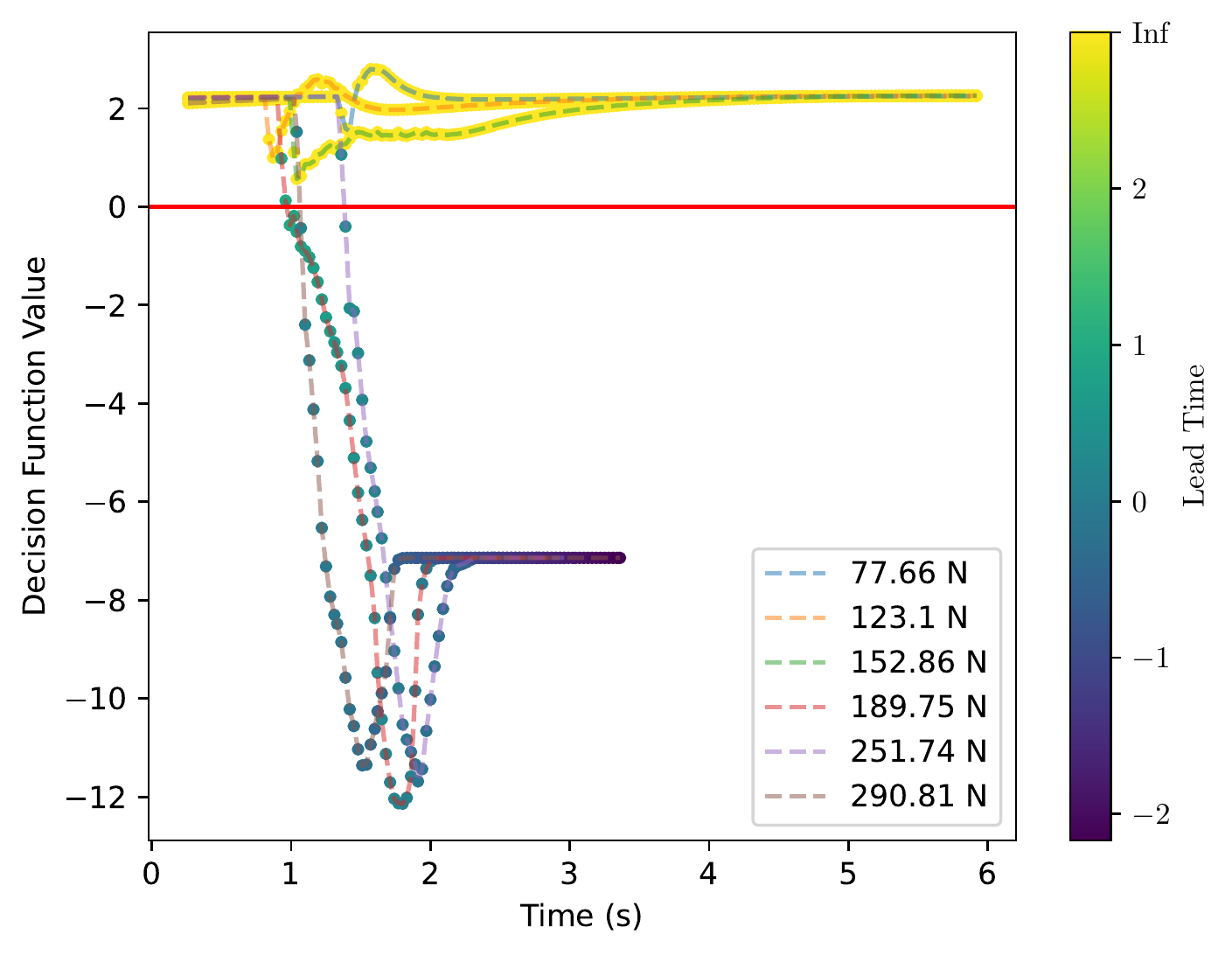}}}    
\caption{Plots displaying the classification results of the nearest-neighbor and SVM classification algorithms for several trajectories. The red solid line is the threshold for the decision function. The dots are the last data point in a window. Positive and negative values for the SVM decision function result in safe and faulty classifications, respectively. On the other hand, values below and above the decision function threshold are classified as safe and faulty for the nearest-neighbor classification algorithm.}
\label{fig:class_results}
\end{figure*}

 From Table \ref{tab:fault_comp_clustering} and \ref{tab:fault_comp_classification} we see that the nearest-neighbor and the SVM classification algorithms perform similarly when trained and evaluated on the abrupt and incipient faults separately. The nearest-neighbor classification algorithm achieves an average lead time of 0.46s and 0.91s, respectively, for the abrupt and incipient faults, while the SVM classification algorithm achieves an average lead time of 0.48s and 0.97s. Because our sampling time is 0.03s, the difference in the performance of both algorithms is 1 and 2 data points for the abrupt and incipient fault respectively.  Fig.~\ref{fig:class_results} displays the classification results for several trajectories.

 When both faults are trained and evaluated together, the SVM classification algorithm achieves an average lead time 0.15s higher compared to the nearest-neighbor classification algorithm. In comparison to its average performance on the abrupt and incipient fault, the SVM classification algorithm achieves an average lead time of 0.08s less when trained on both faults together. Similarly, the nearest-neighbor classification algorithm achieves an average lead time of 0.19s less. As a result, the SVM classification algorithm outperforms the nearest-neighbor classification algorithm when both faults are trained together. However, because both algorithms can achieve lead times higher than the 0.2s, which is the lead time required by reflexive algorithms such as \cite{ref:hohn2009probabilistic} and \cite{ref:wu2021falling}, both algorithms are viable options. As the nearest-neighbor classification algorithm learns the safe/good model, it should be used when faulty data is sparse.

\subsection{Categorizing Faults}
A means to decrease the difference in performance for both algorithms when trained on both faults together vs. separately is to implement a multi-class classification problem. The labels for this multi-class classification can be identified as: abrupt fault safe (AS), abrupt fault fall (AF), incipient fault safe (IS), and incipient fault fall (IF). Using these labels with the one-vs-one or one-vs-rest multi-class classification techniques typically implemented \cite{ref:bishop2006pattern}, results in six and four detectors, respectively. However, as one-vs-rest can result in ambiguities and class imbalances and using one-vs-one can result in ambiguities and higher computational times, we seek a different approach \cite{ref:bishop2006pattern}.

If the problem is decomposed into classifying trajectories first into the incipient versus abrupt categories, and secondly, detecting falls (or not) within these categories of trajectories, the number of detectors needed is only three: a detector for identifying types of trajectories, a second for detecting falls in incipient trajectories, and a third for detecting falls in abrupt trajectories. Furthermore, using this technique resolves the ambiguity problem as the incipient vs. abrupt classifier can be used to determine the operational space (abrupt vs incipient fault). 

To achieve this, we propose using SVM to categorize the trajectories into incipient vs abrupt. The training data for this SVM are taken as the joint velocities, and the labels 1 and -1 are used for the incipient and abrupt faults, respectively. For the training data, the windows in abrupt trajectories before a force is applied and windows uniformly distributed throughout the incipient trajectories are labeled as incipient and only the windows containing the force are labeled as abrupt. The rest of the pre-processing process follows steps similar to those in Section \ref{sec:data_preparation}. 

\begin{table}
        \centering
        \caption{ A comparison of the nearest-neighbor classification algorithm's performance when trained with (1) just the abrupt fault, (2) just the incipient fault, and (3) both faults together. Note that the false positive and negative rates are 0.}
        \tabulinesep=0.5mm
           \begin{tabu}{|c|c|c|c|  }
     \hline\hline
      &\multicolumn{1}{c|}{\makecell{Abrupt Fault\\Only}} & \multicolumn{1}{c|}{\makecell{Incipient Fault\\Only}}& \multicolumn{1}{c|}{\makecell{Both Faults\\Together}} \\
    \hline
           Fold                                                 &\makecell{Average\\Lead Time}         &\makecell{Average\\Lead Time} 
                  &\makecell{Average\\Lead Time}  \\                                             \hline
        1 & 0.48 &  0.93   & 0.49 \\
        \hline
        2    & 0.46  & 0.91  & 0.49 \\
        \hline
        3    & 0.44  & 0.9  & 0.51 \\    
        \hline
        4    & 0.43  & 0.89  & 0.51  \\    
        \hline
        5   & 0.5  & 0.89  & 0.51  \\ 
        \hline
        Average   & 0.46 & 0.91 &  0.50 \\       
    \hline                                               
    \end{tabu}
        \label{tab:fault_comp_clustering}
    \end{table}

\begin{table}
        \centering
        \caption{ A comparison of the SVM classification algorithm's performance when trained with (1) just the abrupt fault, (2) just the incipient fault, and (3) both faults together. Note that the false positive and negative rates are 0. }
        \tabulinesep=0.5mm
           \begin{tabu}{|c|c|c|c|  }
     \hline\hline
     & \multicolumn{1}{c|}{\makecell{Abrupt Fault\\Only}} & \multicolumn{1}{c|}{\makecell{Incipient Fault\\Only}}& \multicolumn{1}{c|}{\makecell{Both Faults\\Together}} \\
    \hline
           Fold                                                 &\makecell{Average\\Lead Time}         &\makecell{Average\\Lead Time} 
                  &\makecell{Average\\Lead Time}  \\                                             \hline
        1 & 0.5 &  1.0   & 0.65 \\
        \hline
        2    & 0.47  & 0.96  & 0.66 \\
        \hline
        3    & 0.49  & 0.98  & 0.66 \\    
        \hline
        4    & 0.44  & 0.97  & 0.65  \\    
        \hline
        5   & 0.51  & 0.96  & 0.63  \\ 
        \hline
        Average   & 0.48 & 0.97 & 0.65 \\       
    \hline                                               
    \end{tabu}
        \label{tab:fault_comp_classification}
    \end{table}

\section{Multi-class Classification Detection Method}
The proposed multi-class classification fall detection method is comprised of three algorithms, one for detecting falls caused by abrupt faults (abrupt fault detector), another for detecting falls caused by incipient faults (incipient fault detector) and a third for identifying the type of fault (fault type identifier). 

The fault type identifier is first trained and evaluated on the training data. Next, the abrupt fault detector and incipient fault detector are trained using the relevant trajectories and the windows of trajectories misclassified by the fault type identifier. The training lead time is determined similarly as in Section \ref{sec:fault_comparison}. 


When detecting potential faults, we run all three algorithms in parallel. As we initially assume that every trajectory has an incipient fault, the output of the incipient fault detector is utilized by default. However, if the fault identifier classifies a trajectory as containing an abrupt fault, we start using the output of the abrupt fault detector. In other words, our null hypothesis is the incipient fault, while our alternative hypothesis is the abrupt fault. As a result, a delay in the fault identifier only results in a delay in the abrupt fault detector. When an abrupt fault is identified, the fault identifier is no longer used to identify the fault type until it is reset. Inherent in our implementation is that only one fault will be encountered per trajectory.

\subsection{Results of Proposed Multi-Class Classification Algorithm Across All Folds}

We train and evaluate the multi-class classification algorithm using the same parameters and metrics as described in Section \ref{sec:fault_comparison}. On average, across all folds, when trained using the features in \eqref{eq:features_main}, the multi-class classification algorithm achieves 0.06s and 0.05s additional average lead time across all folds for the nearest-neighbor and SVM classification algorithms, respectively. This results in an average lead time difference of 0.13s and 0.03s across all folds for the nearest-neighbor and SVM classification algorithms in comparison to their average when trained with the incipient and abrupt faults separately. Note that the SVM fault identifier has a delay of 0.07s or 3 data points, in detecting abrupt faults across all folds. The results are displayed in Table \ref{tab:multi_class_cluster}  and \ref{tab:multi_class_class}.

Even though the multi-class classification algorithm achieved similar results to the binary classification algorithm, an advantage over binary classification is that different features can be used for each detector. Feature selection algorithms such as sequential feature selection can be used to determine the optimal features. For instance, using the features shown in Table \ref{tab:f1_features} derived from scikit-learn's \cite{ref:scikit-learn} sequential forward feature selection results in an average lead time increase of 0.1s over training a binary classification with \eqref{eq:features_main}. However, to truly take advantage of the multi-class classification algorithm more investigation into optimal feature selection is needed to determine whether the additional average lead time gained can overcome the fault identifier delay.

\begin{table}
        \centering
        \caption{ Features derived from scikit-learn's \cite{ref:scikit-learn} sequential forward feature selection}
        \tabulinesep=0.5mm
           \begin{tabu}{|c|c| }
    \hline
    \makecell{Incipient Fault Features}  &\makecell{Abrupt Fault\\Features} \\                                             \hline
                  $\begin{bmatrix}
        \text{\rm knee~angle} \\
        \text{\rm hip~angle} \\
        \text{\rm vel~hip~angle} \\
        (L_{cop} + L_{com})* sgn(p_{com}^x- p_{f_{mid}}^x) 
    \end{bmatrix}$ &            $\begin{bmatrix}
        p_{com}^x\\
        v_{com}^x\\
        p_{com}^x - p_{heel}^x\\
        \text{\rm foot~x} \\
        \text{\rm vel~foot~z} \\
        \text{\rm vel~hip~angle} 
    \end{bmatrix} $\\
    \hline                                               
    \end{tabu}
        \label{tab:f1_features}
    \end{table}

\begin{table}
        \centering
        \caption{ A comparison of the maximum average lead time achieved by the binary nearest-neighbor classification algorithm and the multi-class classification algorithm with nearest-neighbor fault detectors}
        \tabulinesep=0.5mm
           \begin{tabu}{|c|c|c| }
    \hline
            Fold                                                       &\makecell{Multi-class Classification\\Average Lead Time} 
                  &\makecell{Binary Nearest-Neighbor\\Average Lead Time} \\                                             \hline
        1&  0.56 & 0.49  \\
        \hline
        2    & 0.52 & 0.49\\
        \hline
        3    &  0.57 & 0.51 \\    
        \hline
        4    &  0.56 & 0.51  \\    
        \hline
        5   &  0.57 & 0.51 \\ 
        \hline
        Average  &0.56   & 0.5 \\       
    \hline                                               
    \end{tabu}
        \label{tab:multi_class_cluster}
    \end{table}

\begin{table}
        \centering
        \caption{ A comparison of the maximum average lead time achieved by the binary SVM classifier and the multi-class classification algorithm with SVM fault detectors}
        \tabulinesep=0.5mm
           \begin{tabu}{|c|c|c|c|  }
    \hline
           Fold                                                       &\makecell{Multi-class Classification\\Average Lead Time} 
                  &\makecell{Binary Classification\\Average Lead Time} \\                                             \hline
        1&  0.7 & 0.65  \\
        \hline
        2    & 0.73 & 0.66  \\
        \hline
        3    &  0.70 & 0.66\\    
        \hline
        4    &  0.70 & 0.65  \\    
        \hline
        5   &  0.67 & 0.63  \\ 
        \hline
        Average  &0.7   & 0.65 \\       
    \hline                                               
    \end{tabu}
        \label{tab:multi_class_class}
    \end{table}

\section{Conclusion}
The objective of this paper was to design a fall detection algorithm for bipedal robots that is capable of detecting both incipient and abrupt faults while maximizing the lead time and meeting the desired false positive and negative rates. To meet the desired upper bound on the false positive and negative rates, we proposed using training lead time, a subset of lead time, to label the windows in a trajectory. We successfully implemented a nearest-neighbor fall detection classification algorithm, and analyzed and compared its performance to an SVM classification-based algorithm. Using false positive and negative rate, and average lead time as metrics, we found that the nearest-neighbor classification algorithm's performance is comparable to the SVM classifier when trained on abrupt and incipient faults separately. However, it detects falls on average 0.15s (results to 5 data points given our sampling time) slower than the SVM classifier when the faults are trained together. Given that the nearest-neighbor classification algorithm still has average lead time of 0.5s, we conclude that if a sufficient amount of faulty data is not available, the nearest-neighbor classification algorithm can be used to detect abrupt and incipient faults simultaneously.

Even though the SVM classification algorithm outperforms the nearest-neighbor classification algorithm, its leading time when trained on both faults together is slightly lower than its average lead time from both faults separately. As a result, we investigate the use of a multi-class classification algorithm to reduce this difference. We find, that using the same features with the multi-class classification algorithm increases the average lead time slightly. We briefly investigate using the multi-class classification algorithm with different features for the incipient and abrupt faults, and conclude that the multi-class classification algorithm shows promising results. However, more investigation is needed in feature selection and reduction in the delay time of the fault identifier to truly assess the advantage of using a multi-class classification algorithm.








\textbf{Acknowledgements:} This work was supported in part by NSF Award No.~1808051. In addition, M.E. Mungai was supported in part by an NSF Graduate Research Fellowship and a Rackham Merit Fellowship. The authors thank Prof. Maani Ghaffari for his helpful insights. In addition, M.E. Mungai thanks Grant Gibson, Yves Nazon, Isaack Karanja, and Vaibhav Singh for helpful advice. 

\balance
\bibliographystyle{IEEEtran}
\bibliography{references.bib}
\end{document}